\documentclass[Afour,sagev,times]{sagej}

\usepackage{moreverb,url}
\usepackage{makecell,rotating,multirow,diagbox}
\usepackage{hyperref}
\hypersetup{colorlinks,bookmarksopen,bookmarksnumbered,citecolor=red,urlcolor=red}

\newcommand\BibTeX{{\rmfamily B\kern-.05em \textsc{i\kern-.025em b}\kern-.08em
T\kern-.1667em\lower.7ex\hbox{E}\kern-.125emX}}

\def\volumeyear{2019}

\begin{document}

\title{Object Sorting Using a Global Texture-Shape 3D Feature Descriptor}
\author{Zhun Fan\affilnum{1}, Zhongxing Li\affilnum{1}, Benzhang Qiu\affilnum{1}, Wenji Li\affilnum{1}, Jianye Hu\affilnum{1}, Alex Noel Josephraj\affilnum{1} and Heping Chen\affilnum{2}}
%\author{Alistair Smith\affilnum{1} and Hendrik Wittkopf\affilnum{2}}

\affiliation{
  \affilnum{1}The Guangdong Provincial Key Laboratory of Digital Signal and Image Processing, College of Engineering, Shantou University, Shantou 515063, China\\
  \affilnum{2}The Ingram School of Engineering, Texas State University, San Marcos, TX 78666 USA
  }

\corrauth{Zhun Fan,\\
Provincial Key Laboratory of Digital Signal and Image Processing,\\
College of Engineering, Shantou University,\\
Shantou 515063, China.}

\email{zfan@stu.edu.cn}

\begin{abstract}
  Object recognition and grasping plays a key role in robotic systems, especially for the autonomous robots to implement object sorting tasks in a warehouse. 
  In this paper, we present a global texture-shape 3D feature descriptor which can be utilized in a system of object recognition and grasping, and can perform object sorting tasks well. 
  Our proposed descriptor stems from the clustered viewpoint feature histogram (CVFH), 
  which relies on the geometrical information of the whole 3D object surface only, and can not perform well in recognizing the objects with similar geometrical information. 
  Therefore, we extend the CVFH descriptor with texture and color information to generate a new global 3D feature descriptor. 
  The proposed descriptor is evaluated in tasks of recognizing and classifying 3D objects by applying multi-class support vector machines (SVM) in both public 3D image dataset and real scenes. 
  The results of evaluation show that the proposed descriptor achieves a significant better performance for object recognition compared with the original CVFH. 
  Then, the proposed descriptor is applied in our object recognition and grasping system, showing that the proposed descriptor helps the system implement the object recognition, object grasping and object sorting tasks well.
  
  \end{abstract}


\maketitle

\section{Introduction and related work}
With the development of e-commerce, it becomes more and more important for the autonomous robots to execute object sorting task in a warehouse, 
in which they need to locate the target objects, identify what the objects are, and grasp the objects to right places. 
All the information needed for these tasks can be provided by 3D images. 
However, it is still a challenging and difficult task to design a reliable and effective system for object recognition\cite{correll2016analysis}.

It is a common way to utilize the features of the target objects in the 2D image plane for executing object recognition. 
There are several examples of these approaches, including SIFT\cite{lowe2004distinctive}, SURF\cite{bay2008speeded} and ORB\cite{rublee2011orb}. 
Even though these methods can perform well on objects with rich texture, they show bad performance on textureless objects.

In recent years, more and more low-cost, effective 3D sensors are available for object recognition. 
These 3D sensors can obtain both shape and texture information of the target objects, such as the Asus Xtion camera and the 3D (RGB-D) Microsoft Kinect\cite{zhang2012microsoft}. 
As a result, the 3D feature descriptors have become a popular way for object recognition \cite{fan2017combined}. 
The 3D feature descriptors are divided into two types, one is local 3D feature descriptor, and the other is global 3D feature descriptor.

The local 3D feature descriptor relies on the geometrical information of the 3D object surface. 
Each local 3D feature descriptor is computed for a single key point of the 3D object surface, thus each 3D object has more than one local 3D feature descriptor. 
A set of all local 3D feature descriptors can be used for object recognition by means of feature matching. 
Several local 3D feature descriptors are proposed for object recognition, 
like the 3D shape context (3DSC)\cite{frome2004recognizing}, the unique shape context descriptor (USC)\cite{tombari2010unique}, 
the normal aligned radial feature (NARF)\cite{steder2011point}, and the rotational projection statistics (RoPS)\cite{guo2013rotational}. 
These descriptors can perform well on objects containing rich geometrical information for object recognition. 
In addition, they do not need to undergo the stage of segmentation when they are applied in object recognition. 
However, they need a lot of computational resources for computing descriptors and descriptor matching when they are used for object recognition.

\begin{figure*}
  \begin{minipage}{0.2\linewidth}
    \centerline{\includegraphics[width=1\textwidth]{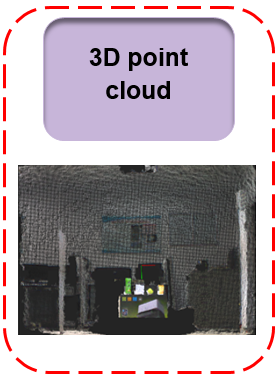}}
    \centerline{(a)}
  \end{minipage}
  \begin{minipage}{0.36\linewidth}
    \centerline{\includegraphics[width=1\textwidth]{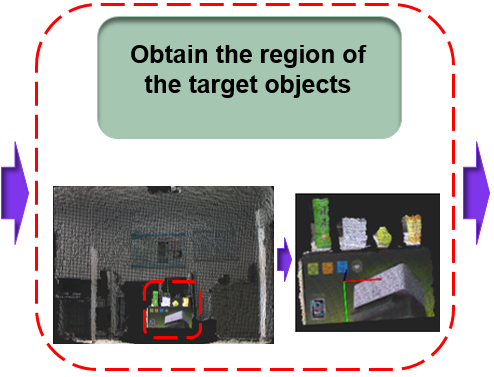}}
    \centerline{(b)}
  \end{minipage}
  \begin{minipage}{0.205\linewidth}
    \centerline{\includegraphics[width=1\textwidth]{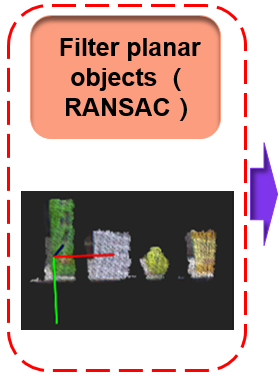}}
    \centerline{(c)}
  \end{minipage}
  \begin{minipage}{0.168\linewidth}
    \centerline{\includegraphics[width=1\textwidth]{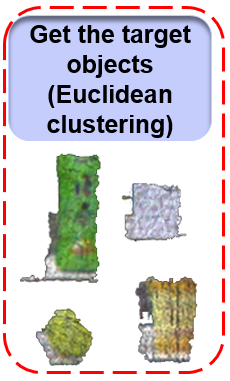}}
    \centerline{(d)}
  \end{minipage}
  \caption{The process of the proposed method for segmentation: 
  (a) Obtain the 3D point cloud image of the real scene. 
  (b) Obtain the regions of the target objects. 
  (c) Filter planar objects by using RANSAC algorithm. 
  (d) Get the target objects by using Euclidean Cluster algorithm.}
  \label{fig_process}
  \end{figure*}

The global 3D feature descriptor relies on the geometrical information of the whole 3D object surface, 
which is computed for a whole 3D object, instead of for a single key point of the 3D object surface. 
Thus every object has only one single global 3D feature descriptor. 
These descriptors can also be used for executing object recognition and object classification by means of feature matching. 
However, they need to extract the target objects from the cluttered scenes before they are applied in object recognition. 
There are several global 3D feature descriptors used for object recognition, 
including the global fast point feature histogram (GFPFH)\cite{rusu2009detecting}, the ensemble of shape functions (ESF)\cite{wohlkinger2011ensemble}, 
the oriented, unique and repeatable clustered viewpoint feature histogram (OUR-CVFH)\cite{aldoma2012our} and the global radius-based surface descriptor (GRSD)\cite{marton2010hierarchical}. 
When these descriptors are used for object recognition, they require less computational resource for computing descriptors and descriptor matching in comparison with the local 3D feature descriptors.

These two kinds of 3D feature descriptors rely on the geometrical information of the 3D object surface, which enables them to obtain a good performance of objects recognition on the object with different shapes. 
But they cannot guarantee performance on the objects with similar shapes. 
Some 3D feature descriptors are based on both shape and texture information, like the Color-SHOT (CSHOT)\cite{tombari2011combined}, 
which is a local 3D feature descriptor which combines the shape information with the color information. 
When these local 3D feature descriptors are used for object recognition, more computational resources are required for computing descriptors and descriptor matching.

In this work, we propose a global texture-shape 3D feature descriptor for object recognition and grasping. 
The major contributions of the paper are: 
(1) A global texture-shape 3D feature descriptor is proposed by extending the clustered viewpoint feature histogram (CVFH)\cite{aldoma2011cad}. 
(2) The recognition performance of the proposed descriptor is evaluated with both public 3D image dataset and real scenes. 
(3) A multi-class support vector machine (SVM)\cite{vapnik1998support} classifier is used for recognizing target objects instead of using feature matching, which can significantly reduce the computing burden.

% The paper is organized as follows. 
% The system of object recognition and grasping using the proposed feature descriptor is described in Section II. 
% The experimental results are presented in Section III. 
% Finally, we make conclusions and discuss future work in Section IV.

\section{System description}
In this section, the method of segmentation is first given. 
Then the global texture-shape 3D feature descriptor is explained in detail. After that, the proposed object recognition and grasping system is presented.

\subsection{The method of segmentation}
Here, we present a method of segmentation as the preprocess before applying the proposed 3D global descriptor to recognize the target objects. 
In this paper, we utilize the 3D sensor Microsoft Kinect camera to capture the point cloud image of a real scene. 
Figure \ref{fig_process} shows the process of the proposed method for segmentation.

The first step of segmentation is obtaining the 3D point cloud image of the real scene, which is captured by Microsoft Kinect camera. 
We need to obtain the region of the target objects by filtering the point cloud image of the real source scene. 
Then we subtract large planar objects from the real scene by utilizing the random sample consensus (RANSAC)\cite{fischler1981random} algorithm. 
In the end, we can obtain each single target object from the rest point cloud image of the real scene by employing the Euclidean Cluster Algorithm\cite{RusuDoctoralDissertation}.

\emph{Obtain the region of the target objects:} 
After obtaining the point cloud of the source scene, foreground and background subtraction is performed to generate another new point cloud of scene. 
Then the method of segmentation filters the information on the left and right side of the new scene point cloud. 
After that, we can obtain the region of the target objects.

\emph{RANSAC algorithm}\cite{fischler1981random}\emph{:} 
First of all, the algorithm randomly selects a set of points from the point cloud image of source scene. 
Then, the RANSAC algorithm find a mathematical model which fits this selected point set, and computes the parameters of this mathematical model. 
Then this algorithm estimates the error value of every single point in the rest point cloud image according to the mathematical model. 
If the error value is lower than a predetermined threshold value, this point can fit the mathematical model and is considered as an inlier. 
Otherwise, this point is considered as an outlier. 
By repeating the previous steps, after all iterations, we can obtain a set of points which contains the largest number of inliers, which is the planar object we need to subtract.

\emph{The algorithm of the Euclidean Cluster}\cite{RusuDoctoralDissertation}\emph{:} 
The algorithm first creates an empty point set ${S_1}$, and randomly adds a point ${m_0}$ of the point cloud image to this point set ${S_1}$ ($S_1=\{m_0\}$). 
This algorithm computes the distance between the point ${m_0}$ and each of its neighboring point ${n_i}$($i=1,2,...,k$). 
After that, adding the point ${n_k}$ to the point set ${S_1}$ ($S_1=\{m_0,n_k\}$) if the value of distance is lower than the predetermined threshold value. 
Then the algorithm randomly selects another point from the point set ${S_1}$,
and calculates the distance between this point and each of its neighboring point. 
It then adds the suitable points with the distance lower than the predetermined threshold value to the the point set ${S_1}$. 
Repeating the steps until all points of the point cloud image have been processed. 
Then we obtain a point set ${S_1}$ which is an individual object we need to extract from the real scene. 
Similarly, we can obtain other individual objects from the real scene by using this algorithm.

\subsection{The design of the global texture-shape 3D feature descriptor}
Combining the shape information with color information is an effective way to design a reliable object recognition system. 
The proposed global 3D feature descriptor extends the clustered viewpoint feature histogram (CVFH)\cite{aldoma2011cad} by integrating color information, and is denoted as Color-CVFH. 
Next, we first give a brief description of the CVFH descriptor.

\emph{CVFH descriptor}\cite{aldoma2011cad}\emph{:} 
The CVFH stems from the viewpoint feature histogram (VFH)\cite{rusu2010fast}. 
The VFH descriptor is a 3D global descriptor which is formed by four different angular distributions of the object surface normals. 
In addition, the VFH consists of two components, including viewpoint direction component and extended fast point feature histograms (FPFH)\cite{rusu2009fast} component. 
Here, the centroid of the whole surface points is denoted by ${p_c}$, and the normal of this centroid ${p_c}$ is named as ${n_c}$ ($||n_c||=1$). 
Also, we set a local reference coordinate frame ($u_i$, $v_i$, $w_i$) for every point $p_i$($i=1,2,...,k$) of the object surface point, then we can obtain the following equations\cite{rusu2010fast}.

  \begin{equation}
  \left\{
\begin{aligned}
&{u_i}={n_c} \\
&{v_i}=\frac{p_i-p_c}{||p_i-p_c||} \times{u_i} \\
&{w_i}={u_i}\times{v_i}
\end{aligned}
\right.
\end{equation}

The norm of every point $p_i$ of the object surface point is name as ${n_i}$, and we can obtain four different normal angular deviations by using the following equations\cite{rusu2010fast}.

  \begin{equation}
  \left\{
\begin{aligned}
&\cos(\alpha_i)={v_i}\cdot{n_i} \\
&\cos(\beta_i)={n_i}\cdot\frac{p_c}{||p_c||}\\
&\cos(\Phi_i)={u_i}\cdot\frac{p_i-p_c}{||p_i-p_c||}\\
&{\theta_i}={atan2(w_i}\cdot{n_i,u_i}\cdot{n_i)}
\end{aligned}
\right.
\end{equation}

In addition, the extended FPFH\cite{rusu2009fast} component is built by utilizing the three different normal angular deviations including $\cos(\alpha_i)$, $\cos(\Phi_i)$ and $\theta_i$, 
and the viewpoint direction component is built by using the normal angular deviation $\cos(\beta_i)$. However, the VFH descriptor can not perform well on occluded objects. 
To solve this problem, the CVFH descriptor computes a VFH histogram for each stable, smooth region of the cluster surface by utilizing the region-growing segmentation, 
rather than one single VFH histogram for the whole cluster surface. 
Additionally, the CVFH descriptor not only contains a viewpoint direction component and an extended FPFH component, but also a new component called the shape distribution component (SDC)\cite{aldoma2011cad}. 
The definition of the SDC is shown as following.

  \begin{equation}
  \left.
\begin{aligned}
&{SDC}=\frac{(p_c-p_i)^2}{\max((p_c-p_i)^2)}\quad\text{where}\;{i=1,2,...,N}
\end{aligned}
\right.
\end{equation}

Where $N$ represents the total number of the whole object surface points. 
The SDC component is described by a histogram of 45 bins, each normal angular deviation of the extended FPFH\cite{rusu2009fast} component is also described by a histogram of 45 bins, 
and the angular deviations of the viewpoint direction component is described by a histogram of 128 bins. 
Therefore, the CVFH descriptor is described by a histogram of 308 bins.

\emph{Color-CVFH descriptor:} 
In this paper, we design a global texture-shape 3D feature descriptor named as Color-CVFH which contains two components, a global color histogram and a CVFH histogram. 
We can obtain the point cloud image of real scene by using the RGB-D sensor Microsoft Kinect, which not only contains the shape information, but also the color information. 
Also, we can acquire the point cloud of each individual target object from the real scene by utilizing the proposed method of segmentation. 
Thus we can get the color information of the target object, which is now described in RGB color space. 
However, in most circumstance, the color information which is described in HSV (Hue, Saturation, Value) color space outperforms the color information in RGB color space for perception. 
Therefore, in this work, we transform the color information from RGB color space to HSV color space. 
The hue dimension is a primary component in HSV color space for perception, and is divided into a histogram of 90 bins. 
Also, the saturation dimension and value dimension are both divided into a histogram of 51 bins. After that, we can obtain a global color histogram feature which contains 192 histogram bins. 
It is noted that one single object has only a global color histogram. 
Here, we denote the global color histogram feature of the object $O_i$ as $M_i$, the CVFH histogram feature of the object $O_i$ as $N_i$, and the Color-CVFH descriptor of the object $O_i$ as $L_i$. 
Mathematically, the definition of the Color-CVFH descriptor $L_i$ is shown as following.

  \begin{equation}
  \left.
\begin{aligned}
&{L_i}={M_i}\cup{N_i}
\end{aligned}
\right.
\end{equation}

\begin{figure*}[htbp]
  \centering
  \includegraphics[width=0.98\textwidth]{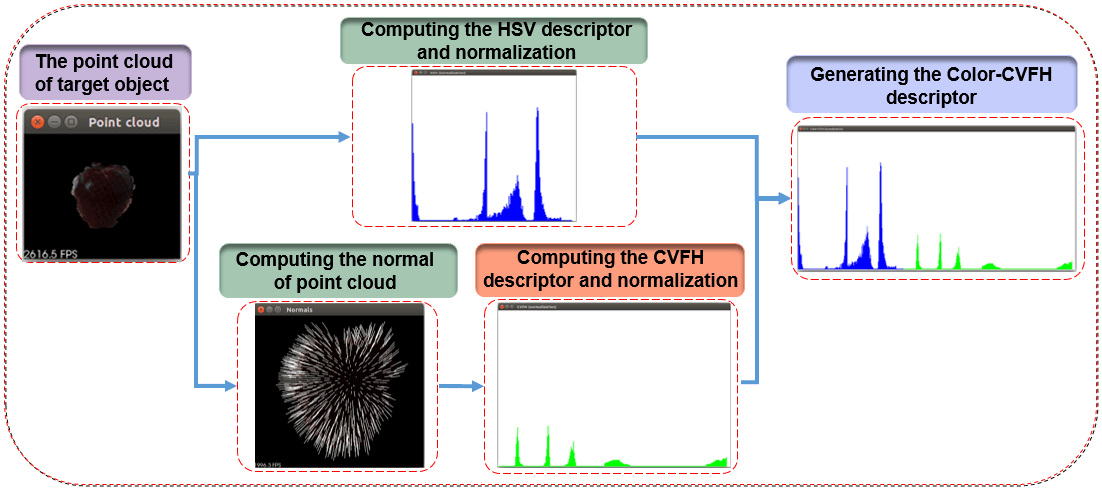}
  \caption{The details of generating Color-CVFH descriptor}\label{fig_c_cvfh}
\end{figure*}

The equation (4) shows that the Color-CVFH descriptor $L_i$ consists of two different global histograms. 
The first part of the Color-CVFH is a global color histogram $M_i$, followed by the CVFH histogram $N_i$ as the second part. 
Obviously, the Color-CVFH descriptor is a global 3D feature descriptor which is built by a histogram of 500 bins (HSV: 192 bins, CVFH: 308 bins). 
Because the Color-CVFH descriptor makes use of both color and geometrical information, it can be conveniently utilized to train a multi-class SVM classifier for recognizing and classifying the target objects.
Therefore, the details of generating the proposed descriptor can be shown in Figure \ref{fig_c_cvfh}.

\subsection{The design of object recognition and grasping system}
\begin{figure}[htbp]
  \centering
  \includegraphics[width=0.48\textwidth]{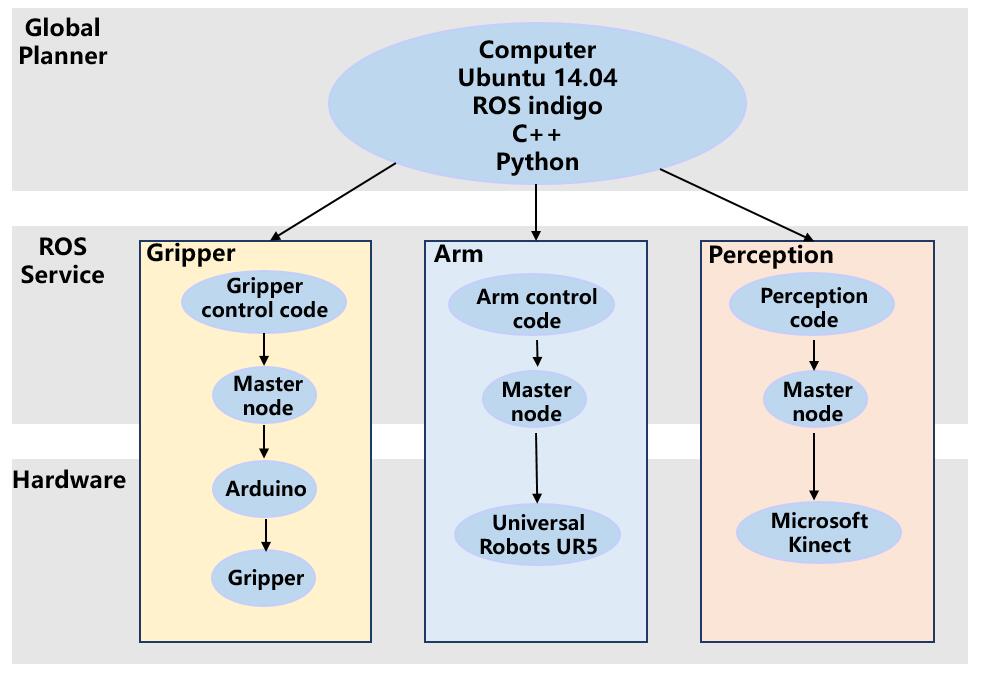}
  \caption{The architecture of the object recognition and grasping system}\label{fig_architecture}
\end{figure}

\begin{figure}[htbp]
  \centering
  \includegraphics[width=0.48\textwidth]{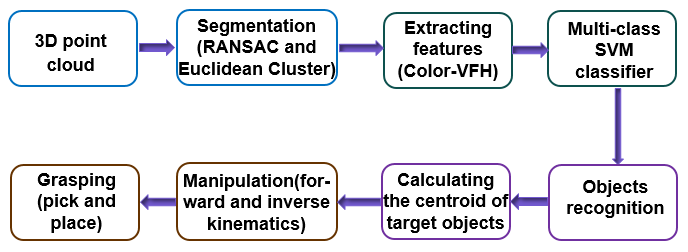}
  \caption{The functional architecture of the object recognition and grasping system}\label{fig_func_architecture}
\end{figure}

In this section, we present a system of object recognition and grasping, which is based on the proposed descriptor Color-CVFH. 
The proposed system is able to execute the object recognition, object grasping and object sorting tasks in a sequence. 
The communication between different processes in this proposed system is handled by the Robot Operating System (ROS)\cite{quigley2009ros}. 
The architecture and functional architecture of the proposed system are shown in Figure \ref{fig_architecture} and Figure \ref{fig_func_architecture} respectively.

The Figure \ref{fig_func_architecture} shows that the first step of the object recognition and grasping system is capturing the point cloud image of the real scene by the 3D sensor Microsoft Kinect camera. 
Next, each individual target object is extracted from the real scene by utilizing the RANSAC\cite{fischler1981random} algorithm and Euclidean Cluster\cite{RusuDoctoralDissertation} algorithm. 
After the target objects are extracted, the proposed feature descriptor Color-CVFH of these target objects can be computed, and used to train the multi-class SVM classifier.
The classifier is applied to identify the category, and acquire the position of each individual target object. 
In addition, the centroid of each target object is considered as the position for the robotic manipulator to grasp. 
Then after obtaining the forward kinematics and inverse kinematics solutions, this system can control the end effector of the robotic manipulator to approach the target object. 
Then, the system execute the object grasping according to the centroid of the target object. 
In the end, the robotic manipulator can grasp the target objects, and place these target objects to their specified positions according to their categories. 
In this way, the object recognition and grasping system completes the sequence of tasks of object recognition, grasping and sorting.

\begin{figure}[htbp]
  \centering
  \includegraphics[width=0.48\textwidth]{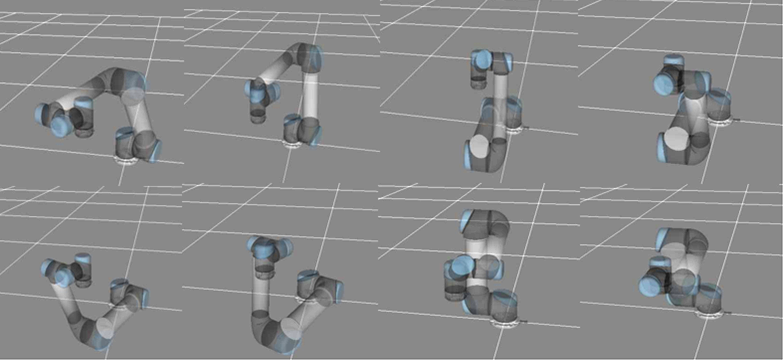}
  \caption{An example of 8 inverse kinematics solutions by using geometric method.}\label{fig_ik}
\end{figure}

\emph{Forward kinematics and inverse kinematics:} 
In this paper, the object recognition and grasping system is run on a Universal Robots UR5, which is a manipulator with 6 revolute joints. 
In the object recognition and grasping system, the forward kinematics of the Universal Robots UR5 is calculated by the Denavit-Hartenberg (D-H) parameters\cite{denavit1955kinematic}. 
Here we calculate the inverse kinematics solution of the Universal Robots UR5 by using the the geometric method\cite{hawkins2013analytic}. 
By leveraging the geometric method, we can obtain 8 inverse kinematics solutions. 
An example of 8 inverse kinematics solutions by using geometric method is shown in Figure \ref{fig_ik}. 
This geometric method not only saves the computational time, but also acquires high accuracy of the inverse kinematics solutions.

\section{Experiments and results}
We begin this section by evaluating the proposed descriptor Color-CVFH in a public RGB-D image dataset\cite{lai2011large}, which is constructed by Washington University. 
We evaluate the recognition performance of the proposed descriptor Color-CVFH in this public dataset at two levels, including category level and instance level recognition. 
Besides, the Color-CVFH descriptor is evaluated by real scenes (150 different scenes sampled in total).

In this work, the recognition performance of the Color-CVFH descriptor and the CVFH descriptor are evaluated by four metrics, including recall, precision, accuracy and F1\_score. 
The four metrics are defined as following.

\begin{equation}
\left.
\begin{aligned}
&{Recall}=\frac{TP}{TP+FN}
\end{aligned}
\right.
\end{equation}

\begin{equation}
\left.
\begin{aligned}
&{Precision}=\frac{TP}{TP+FP}
\end{aligned}
\right.
\end{equation}

\begin{equation}
\left.
\begin{aligned}
&{Accuracy}=\frac{TP+TN}{TP+TN+FP+FN}
\end{aligned}
\right.
\end{equation}

\begin{equation}
\left.
\begin{aligned}
&{F1\_score}=\frac{{2}\cdot{Precision}\cdot{Recall}}{Precision+Recall}
\end{aligned}
\right.
\end{equation}

Where TP represents the number of true positives, FP represents the number of false positives, TN represents the number of true negatives, and FN represents the number of false negatives.

In the end, we present the proposed system of object recognition and grasping, and test it by using the proposed descriptor Color-CVFH to execute the object recognition, grasping and sorting tasks in a sequence.

\subsection{Evaluated by a public dataset at category level recognition}
\begin{figure}[htbp]
  \centering
  \includegraphics[width=0.48\textwidth]{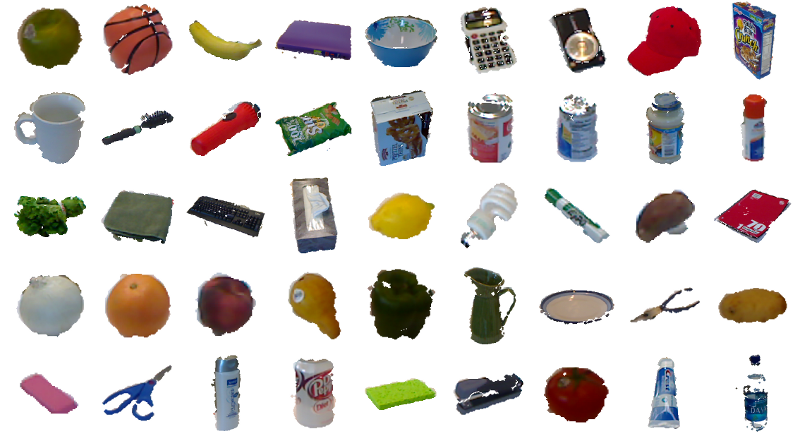}
  \caption{The samples of the public RGB-D image dataset}\label{fig_data_sample}
\end{figure}

In this experiment, the public RGB-D image dataset created by Washington University\cite{lai2011large}  is utilized for evaluating the recognition performance of the Color-CVFH descriptor. 
This public RGB-D image dataset contains 300 common household objects in 51 categories, and each object (instance) is captured by 3 different video sequences. 
Some samples of the public RGB-D image dataset are shown in Figure \ref{fig_data_sample}.

\begin{figure}[htbp]
  \centering
  \includegraphics[width=0.46\textwidth]{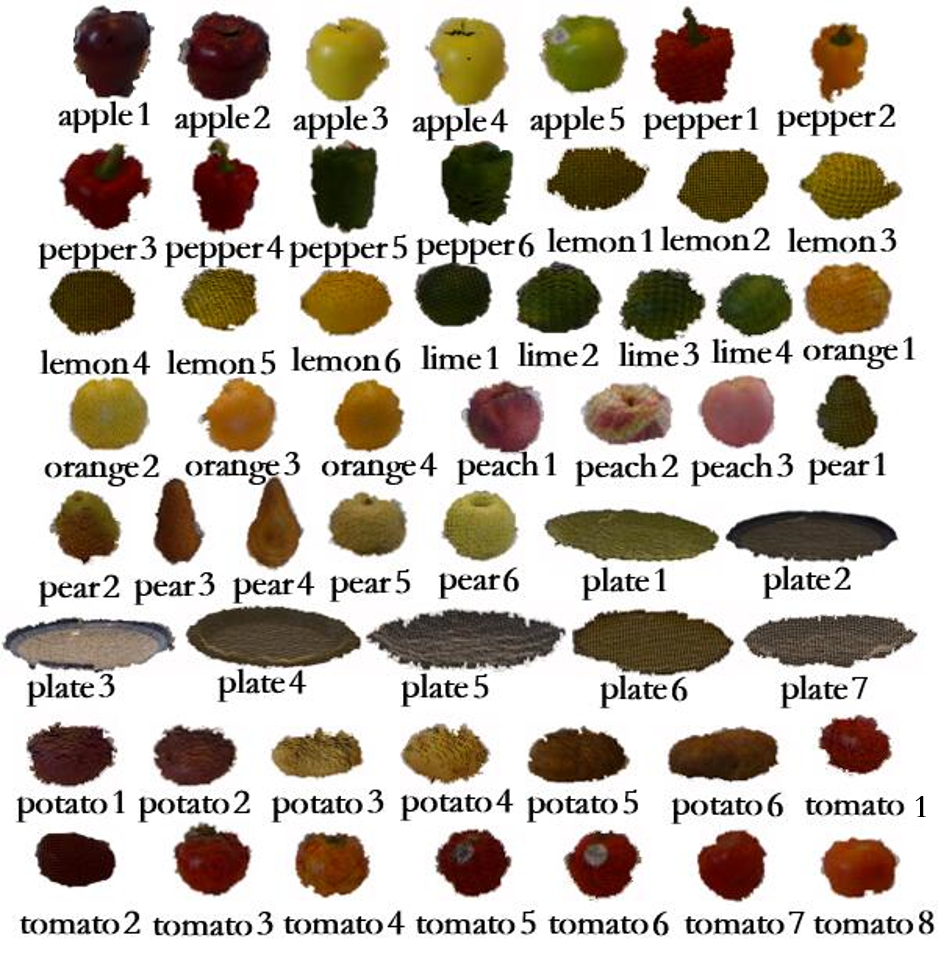}
  \caption{One view of the ten categories used in the experiment.}\label{fig_train_data}
\end{figure}

In this public dataset, each category contains several different objects (instances). 
The category level recognition aims at classifying category that previously unseen objects belong to, such as recognizing the object apple2 (as shown in Figure \ref{fig_experiment_data}) belonging to the category of apple. 
In this experiment, we select ten categories of the objects from this public dataset, which are apple, pepper, lemon, lime, orange, peach, pear, plate, potato and tomato. 
These ten categories of objects are shown in Figure \ref{fig_train_data}. 

This experiment is conducted for evaluating and comparing the recognition performance of the Color-CVFH descriptor and the CVFH descriptor by using a multi-class SVM classifier. 
From these ten categories of objects, we select 26731 different point clouds in total (each object consists of 3 different video sequences) for training the multi-class SVM classifier. 
We select the rest of 8835 different point clouds in total (each object consists of 3 different video sequences) for testing. 
The results of the experiment are shown in Table \ref{tab_1}, Table \ref{tab_2} and Table \ref{tab_3}.

\begin{table*}[htbp]
  \renewcommand{\arraystretch}{1.5}
    \centering
    \caption{The results of the evaluation for CVFH and Color-CVFH(normalization) by using the public dataset at category level recognition}
    \label{tab_1}
      \begin{tabular}{|c|c|c|c|c|c|c|c|c|}
      \hline
      \multirow{2}{*}{}&
      \multicolumn{4}{c|}{CVFH}&\multicolumn{4}{c|}{Color-CVFH (normalization)}\cr\cline{2-9}
      &Recall&Precision&Accuracy&F1\_score&Recall&Precision&Accuracy&F1\_score\cr
      \hline
      apple vs. all&53.90\%&40.25\%&89.01\%&0.4609&{\bf 100\%}&{\bf 100\%}&{\bf 100\%}&{\bf 1}\cr\hline
      pepper vs. all&81.47\%&76.42\%&95.36\%&0.7887&{\bf 100\%}&{\bf 100\%}&{\bf 100\%}&{\bf 1}\cr\hline
      lemon vs. all&64.69\%&58.29\%&91.42\%&0.6133&{\bf 100\%}&{\bf 100\%}&{\bf 100\%}&{\bf 1}\cr\hline
      lime vs. all&73.92\%&52.44\%&93.41\%&0.6135&{\bf 100\%}&{\bf 100\%}&{\bf 100\%}&{\bf 1}\cr\hline
      orange vs. all&70.55\%&85.52\%&96.71\%&0.7732&{\bf 100\%}&{\bf 100\%}&{\bf 100\%}&{\bf 1}\cr\hline
      peach vs. all&34.04\%&23.95\%&89.81\%&0.2812&{\bf 100\%}&{\bf 100\%}&{\bf 100\%}&{\bf 1}\cr\hline
      pear vs. all&14.09\%&32.10\%&86.34\%&0.1959&{\bf 100\%}&{\bf 99.90\%}&{\bf 99.99\%}&{\bf 0.9995}\cr\hline
      plate vs. all&100\%&100\%&100\%&1&100\%&100\%&100\%&1\cr\hline
      potato vs. all&44.99\%&29.19\%&82.94\%&0.3541&{\bf 99.89\%}&{\bf 100\%}&{\bf 99.99\%}&{\bf 0.9995}\cr\hline
      tomato vs. all&15.08\%&32.03\%&82.80\%&0.2050&{\bf 100\%}&{\bf 100\%}&{\bf 100\%}&{\bf 1}\cr\hline
      \hline
      \end{tabular}
  \end{table*}

  \begin{table*}[htbp]
  \renewcommand{\arraystretch}{1.5}
    \centering
    \caption{The results of prediction for CVFH by using the public dataset at category level recognition}
    \label{tab_2}
    \resizebox{\textwidth}{35mm}{
      \begin{tabular}{|c|c|c|c|c|c|c|c|c|c|c|}
      \hline
      \diagbox{Actual class}{Prediction class}&apple& pepper&lemon&lime&orange&peach&pear&plate&potato&tomato\cr
      \hline
      apple&{\bf53.90\%}&0&0.52\%&0&1.04\%&33.25\%&1.04\%&0&0.39\%&9.87\%\cr\hline
      pepper &0.32\%&{\bf81.47\%}&0&0&0&0.32\%&1.81\%&0&16.08\%&0\cr\hline
      lemon&1.72\%&0&{\bf64.69\%}&14.96\%&0&0.54\%&2.15\%&0&5.06\%&10.87\%\cr\hline
      lime&1.12\%&0&12.32\%&{\bf73.92\%}&0&0.48\%&0.32\%&0&5.06\%&10.87\%\cr\hline
      orange&6.83\%&0.14\%&16.64\%&0&{\bf70.55\%}&5.83\%&0&0&0&0\cr\hline
      peach&10.06\%&0.19\%&15.28\%&4.06\%&0&{\bf34.04\%}&13.15\%&0&18.57\%&4.64\%\cr\hline
      pear&18.12\%&7.57\%&5.75\%&0&3.26\%&17.35\%&{\bf14.09\%}&0&21.19\%&12.66\%\cr\hline
      plate &0&0&0&0&0&0&0&{\bf100\%}&0&0\cr\hline
      potato &9.48\%&16.67\%&0.65\%&9.59\%&0.87\%&2.18\%&10.46\%&0&{\bf44.99\%}&5.12\%\cr\hline
      tomato&16.46\%&0.15\%&6.69\%&13.15\%&2.62\%&3.85\%&7.69\%&0&34.31\%&{\bf15.08\%}\cr
      \hline
      \end{tabular}}
  \end{table*}
  
  \begin{table*}[htbp]
  \renewcommand{\arraystretch}{1.5}
    \centering
    \caption{The results of the evaluation for Color-CVFH and Color-CVFH(normalization) by using the public dataset at category level recognition}
    \label{tab_3}
      \begin{tabular}{|c|c|c|c|c|c|c|c|c|}
      \hline
      \multirow{2}{*}{}&
      \multicolumn{4}{c|}{Color-CVFH}&\multicolumn{4}{c|}{Color-CVFH (normalization)}\cr\cline{2-9}
      &Recall&Precision&Accuracy&F1\_score&Recall&Precision&Accuracy&F1\_score\cr
      \hline
      apple vs. all&100\%&100\%&100\%&1&100\%&100\%&100\%&1\cr\hline
      pepper vs. all&100\%&100\%&100\%&1&100\%&100\%&100\%&1\cr\hline
      lemon vs. all&100\%&100\%&100\%&1&100\%&100\%&100\%&1\cr\hline
      lime vs. all&100\%&100\%&100\%&1&100\%&100\%&100\%&1\cr\hline
      orange vs. all&100\%&100\%&100\%&1&100\%&100\%&100\%&1\cr\hline
      peach vs. all&100\%&100\%&100\%&1&100\%&100\%&100\%&1\cr\hline
      pear vs. all&96.36\%&92.63\%&98.66\%&0.9445&{\bf 100\%}&{\bf 99.90\%}&{\bf 99.99\%}&{\bf 0.9995}\cr\hline
      plate vs. all&100\%&100\%&100\%&1&100\%&100\%&100\%&1\cr\hline
      potato vs. all&90.85\%&98.12\%&98.87\%&0.9434&{\bf 99.89\%}&{\bf 100\%}&{\bf 99.99\%}&{\bf 0.9995}\cr\hline
      tomato vs. all&100\%&100\%&100\%&1&100\%&100\%&100\%&1\cr\hline
      \hline
      \end{tabular}
  \end{table*}

  From Table \ref{tab_1} and Table \ref{tab_2}, we can see that the classifier trained by the CVFH descriptor has a poor performance compared to the classifier trained by the Color-CVFH descriptor with normalization. 
  It can be further observed from Table \ref{tab_2} that for the classifier trained by the CVFH descriptor, about one third of apple are considered as peach, thus the classifier gets a poor recall (${53.90}\%$) for the category of apple. 
  In addition, this classifier is much more likely to consider the pear and potato as the same category. 
  Over one fifth of pears are recognized as potatoes, and about one tenth of potatoes are considered as pears, 
  thus the classifier has poor performance for these two categories in recall (pear: ${14.09}\%$, potato: ${44.99}\%$) and precision (pear: ${32.10}\%$, potato: ${29.19}\%$). 
  Over one third of tomatoes are recognized as potatoes, so that the classifier has a poor recall (${15.08}\%$) for the category of tomato. 
  However the classifier gets perfect results for the category of plate in the metrics of recall, precision, accuracy and F1\_score. 
  This is because the CVFH descriptor is based on the shape information, and the category of plate has a distinctive shape from other categories in this experiment.
  
  The proposed descriptor Color-CVFH(normalization) can take advantage of both shape and color information of the objects. 
  Thus the classifier trained by the descriptor Color-CVFH can not only perform well on the target objects with different shapes, 
  but also on the objects with similar shapes. 
  From Table \ref{tab_1}, it can be seen that the classifier can classify each target object at category level correctly. 
  Even though apple, pepper, peach and tomato have similar shapes, they are still classified correctly at category level recognition.
  
  From Table \ref{tab_3}, it is noticeable that the classifier trained by the Color-CVFH descriptor with normalization outperforms the classifier trained by the Color-CVFH descriptor without normalization in the metrics of recall, precision, accuracy and F1\_score. 
  The Color-CVFH descriptor with normalization helps the classifier improve its performance for object recognition, such as, the precision of pear increase from ${92.63}\%$ to ${99.90}\%$, and the recall of potato increase from ${90.85}\%$ to ${99.89}\%$.
  
  \subsection{Evaluated by a public dataset at instance level recognition}
\begin{figure}[htbp]
    \centering
    \includegraphics[width=0.48\textwidth]{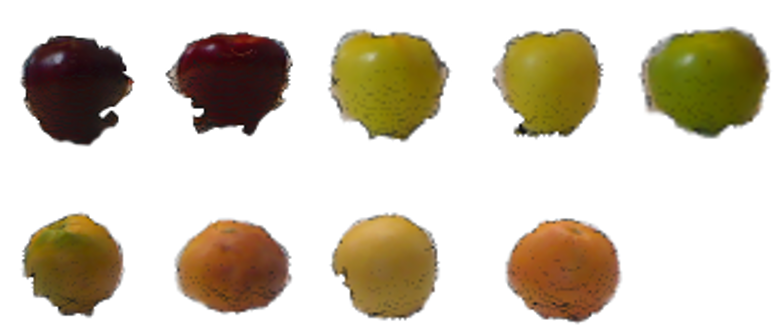}
    \caption{One view of each of the nine objects (instances) used in the experiment. For the first row, left to right: apple1, apple2, apple3, apple4, apple5; For the second row, left to right: orange1, orange2, orange3, orange4.}
    \label{fig_experiment_data}
  \end{figure}

  In this work, we evaluate the recognition performance of the Color-CVFH descriptor and the CVFH descriptor in a public dataset at instance level recognition. 
  The instance level recognition aims at classifying an object and deciding whether it is the same instance that has been previously seen. 
  For example, the object apple1 (as shown in Figure \ref{fig_experiment_data}) is not same as apple2, though these two instances both belong to the category of apple. 
  In this experiment, we select all instances from the categories of the apple and the orange, including apple1, apple2, apple3, apple4, apple5, orange1, orange2, orange3 and orange4. 
  These nine instances are shown in Figure \ref{fig_experiment_data}, from which we can see that apple1 and apple2 have similar shapes and colors, and apple3 and apple4 look pretty much the same object. 
  Orange1, orange2 and orange4 have similar shapes and colors.
  
  For the experiment of instance level recognition, we choose the scenario of alternating contiguous frames\cite{lai2011large}: 
  firstly, we divide each video sequence into three contiguous sequences of equal length. 
  Because each object originally has three different video sequences, each object now has nine video sequences. 
  Then we randomly select seven of these nine video sequences from each object for training multi-class SVM classifier, and select the remaining two for testing.
  
  Here, we select 4642 different point clouds in total for training, and select 1290 different point clouds in total for testing. 
  This experiment is used for evaluating and comparing the recognition performance of the Color-CVFH descriptor and the CVFH descriptor. 
  The results of the experiment are shown in Table \ref{tab_4}, Table \ref{tab_5} and Table \ref{tab_6}.
\begin{table*}[htbp]
  \renewcommand{\arraystretch}{1.5}
  \centering
  \caption{The results of the evaluation for CVFH and Color-CVFH(normalization) by using the public dataset at instance level recognition}
  \label{tab_4}
    \begin{tabular}{|c|c|c|c|c|c|c|c|c|}
    \hline
    \multirow{2}{*}{}&
    \multicolumn{4}{c|}{CVFH}&\multicolumn{4}{c|}{Color-CVFH (normalization)}\cr\cline{2-9}
    &Recall&Precision&Accuracy&F1\_score&Recall&Precision&Accuracy&F1\_score\cr
    \hline
    apple1 vs. all&70.59\%&28.92\%&78.60\%&0.4103&{\bf 90.44\%}&{\bf 100\%}&{\bf 98.99\%}&{\bf 0.9498}\cr\hline
    apple2 vs. all&2.16\%&27.27\%&88.84\%&0.04&{\bf 100\%}&{\bf 91.45\%}&{\bf 98.99\%}&{\bf 0.9553}\cr\hline
    apple3 vs. all&18.80\%&62.50\%&90.47\%&0.289&{\bf 99.25\%}&{\bf 71.35\%}&{\bf 95.81\%}&{\bf 0.8302}\cr\hline
    apple4 vs. all&8.76\%&52.17\%&89.46\%&0.15&{\bf 61.31\%}&{\bf 98.82\%}&{\bf 95.81\%}&{\bf 0.7568}\cr\hline
    apple5 vs. all&25.68\%&20.54\%&80.08\%&0.2282&{\bf 100\%}&{\bf 100\%}&{\bf 100\%}&{\bf 1}\cr\hline
    orange1 vs. all&83.33\%&38.82\%&82.79\%&0.5297&{\bf 100\%}&{\bf 100\%}&{\bf 100\%}&{\bf 1}\cr\hline
    orange2 vs. all&28.19\%&25.61\%&82.25\%&0.2684&{\bf 100\%}&{\bf 100\%}&{\bf 100\%}&{\bf 1}\cr\hline
    orange3 vs. all&56.46\%&61.48\%&91.01\%&0.5887&{\bf 100\%}&{\bf 100\%}&{\bf 100\%}&{\bf 1}\cr\hline
    orange4 vs. all&51.66\%&100\%&94.34\%&0.6812&{\bf 100\%}&100\%&{\bf 100\%}&{\bf 1}\cr\hline
    \hline
    \end{tabular}
\end{table*}

\begin{table*}[htbp]
  \renewcommand{\arraystretch}{1.5}
  \centering
  \caption{The results of prediction for CVFH by using the public dataset at instance level recognition}
  \label{tab_5}
  \resizebox{\textwidth}{35mm}{
    \begin{tabular}{|c|c|c|c|c|c|c|c|c|c|}
    \hline
    \diagbox{Actual class}{Prediction class}&apple1&apple2&apple3&apple4&apple5&orange1&orange2&orange3&orange4\cr
    \hline
    apple1&{\bf70.59\%}&3.68\%&2.94\%&1.47\%&1.47\%&1.47\%&17.65\%&0.74\%&0\cr\hline
    apple2 &66.91\%&{\bf2.16\%}&0&4.32\%&11.51\%&15.11\%&0&0&0\cr\hline
    apple3&34.59\%&1.50\%&{\bf18.80\%}&2.26\%&42.86\%&0&0&0&0\cr\hline
    apple4&46.72\%&0.73\%&0&{\bf8.76\%}&33.58\%&10.22\%&0&0&0\cr\hline
    apple5&20.27\%&0&0&0&{\bf25.68\%}&54.05\%&0&0&0\cr\hline
    orange1&0&0&6.00\%&0&10.00\%&{\bf83.33\%}&0&0.67\%&0\cr\hline
    orange2&2.01\%&0&1.34\%&0&7.38\%&53.69\%&{\bf28.19\%}&7.38\%&0\cr\hline
    orange3 &0&0&0&0&0&0&43.54\%&{\bf56.46\%}&0\cr\hline
    orange4 &0&0&0&0&0&0&22.52\%&25.83\%&{\bf51.66\%}\cr
    \hline
    \end{tabular}}
\end{table*}

\begin{table*}[htbp]
  \renewcommand{\arraystretch}{1.5}
  \centering
  \caption{The results of the evaluation for Color-CVFH and Color-CVFH(normalization) by using the public dataset at instance level recognition}
  \label{tab_6}
    \begin{tabular}{|c|c|c|c|c|c|c|c|c|}
    \hline
    \multirow{2}{*}{}&
    \multicolumn{4}{c|}{Color-CVFH}&\multicolumn{4}{c|}{Color-CVFH (normalization)}\cr\cline{2-9}
    &Recall&Precision&Accuracy&F1\_score&Recall&Precision&Accuracy&F1\_score\cr
    \hline
    apple1 vs. all&100\%&100\%&100\%&1&90.44\%&100\%&98.99\%&0.9498\cr\hline
    apple2 vs. all&100\%&100\%&100\%&1&100\%&91.45\%&98.99\%&0.9553\cr\hline
    apple3 vs. all&43.61\%&89.23\%&93.64\%&0.5859&{\bf 99.25\%}&71.35\%&{\bf 95.81\%}&{\bf 0.8302}\cr\hline
    apple4 vs. all&47.45\%&87.84\%&93.72\%&0.6161&{\bf 61.31\%}&{\bf 98.82\%}&{\bf 95.81\%}&{\bf 0.7568}\cr\hline
    apple5 vs. all&100\%&52.48\%&89.61\%&0.6884&100\%&{\bf 100\%}&{\bf 100\%}&{\bf 1}\cr\hline
    orange1 vs. all&76.00\%&40.43\%&84.19\%&0.5278&{\bf 100\%}&100\%&{\bf 100\%}&{\bf 1}\cr\hline
    orange2 vs. all&97.99\%&100\%&99.77\%&0.9898&{\bf 100\%}&100\%&{\bf 100\%}&{\bf 1}\cr\hline
    orange3 vs. all&43.54\%&64.00\%&90.78\%&0.5182&{\bf 100\%}&{\bf 100\%}&{\bf 100\%}&{\bf 1}\cr\hline
    orange4 vs. all&43.71\%&100\%&93.41\%&0.6083&{\bf 100\%}&100\%&{\bf 100\%}&{\bf 1}\cr\hline
    \hline
    \end{tabular}
\end{table*}

From Table \ref{tab_4} and Table \ref{tab_5}, it can be seen that the classifier trained by the Color-CVFH descriptor with normalization outperforms the classifier trained by the CVFH descriptor in terms of recall, precision, accuracy and F1\_score. 
For the classifier trained by the CVFH(normalization) descriptor, the instances from the category of apple are much more likely to be considered as the object apple1. 
Over half of apple2, about one third of apple3, nearly half of apple4 and over one fifth of apple5 are considered as apple1. 
Thus the classifier has a poor performance for the apple1 in precision ($28.92\%$) and the apple2 in recall ($2.16\%$). 
Over two fifths of apple3 and over one third of apple4 are considered as apple5. 
Thus the classifier has a poor performance for the apple3 in recall ($18.80\%$), the apple4 in recall ($8.76\%$) and the apple5 in precision ($25.68\%$). 
More than half of apple5 and orange2 are considered as orange1, and over two fifths of orange3 are considered as orange2. 
Thus the classifier does not have good performance on orange2 in recall ($28.19\%$), orange3 in recall ($56.46\%$), orange1 in precision ($38.82\%$) and orange2 in precision ($25.61\%$).

By combining the shape and color information, the classifier trained by the descriptor Color-CVFH can perform well in this experiment. 
It can be observed from the Table \ref{tab_5} that most of the target objects are classified correctly at instance level. 
Even though the instances belonging to the category of the orange look similar to each other in both shape and color, they can be classified correctly. 
The classifier acquires excellent performance in terms of recall, precision and F1\_score. 
However, because the apple3 and apple4 look pretty much the same object, this classifier makes some mistakes when recognizing the apple3 and apple4.

From Table \ref{tab_6}, it is noticeable that the classifier trained by the Color-CVFH descriptor with normalization outperforms the classifier trained by the Color-CVFH descriptor without normalization in the metrics of recall, precision, accuracy and F1\_score. 
Obviously, the Color-CVFH descriptor with normalization helps the classifier improve its performance for object recognition. 
For example, the recall of apple3 increases from ${43.61}\%$ to ${99.25}\%$, and the precision of apple5 increases from ${52.48}\%$ to ${100}\%$.

\subsection{Performance evaluation with the real scenes}

\begin{figure}[htbp]
  \centering
  \includegraphics[width=0.25\textwidth]{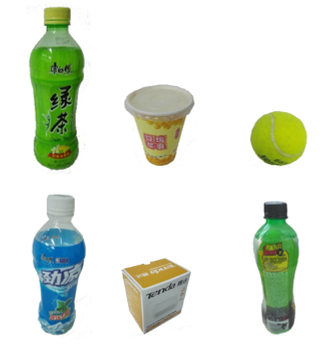}
  \caption{One view of the six chosen objects used in the experiment. For the first row, left to right: bottle1, cup, ball; For the second row, left to right: bottle2, box, bottle3.}
  \label{fig_grasp_object}
\end{figure}

\begin{figure}[htbp]
  \centering
  \includegraphics[width=0.25\textwidth]{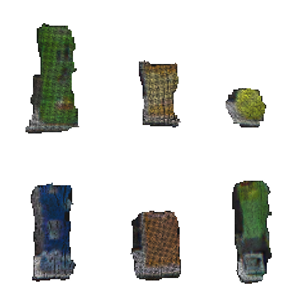}
  \caption{The point cloud of the six chosen objects used in the experiment (One view). For the first row, left to right: bottle1, cup, ball; For the second row, left to right: bottle2, box, bottle3.}
  \label{fig_grasp_object_3d}
\end{figure}

In this work, we evaluate the recognition performance of the Color-CVFH descriptor and the CVFH descriptor in real scenes. 
The objects used for this experiment are bottle1, cup, ball, bottle2, bottle3 and box. These six objects are shown in Figure \ref{fig_grasp_object}, and  Figure \ref{fig_grasp_object_3d} shows the point clouds of them. 
The six objects are further divided into four categories, which are bottle, cup, ball and others. 
The object bottle1, object cup and object ball belong to the category of  bottle1, the category of cup and the category of ball respectively, and the object bottle2, object bottle3 and object box all belong to the category of others. 
The category of others contains 45 different point clouds, and each of the other three categories are built by 100 different point clouds. All these point clouds are acquired by the 3D sensor Microsoft Kinect camera on multiple views. 
It can be observed from Figure \ref{fig_grasp_object} and Figure \ref{fig_grasp_object_3d} that the bottle1 and bottle2 have almost the same shape, but they have different colors. 
However, the bottle1 is similar to bottle3 both in shape and color. The cup and the box share some similarities in shape and color from a certain angle (as shown in Figure \ref{fig_grasp_object_3d}), 
while the ball has distinctive shape in this experiment. 
All these point clouds are used for training the multi-class SVM classifiers to evaluate the recognition performance of the Color-CVFH descriptor and the CVFH descriptor.

\begin{figure*}
  \begin{minipage}{0.33\linewidth}
    \centerline{\includegraphics[width=1\textwidth]{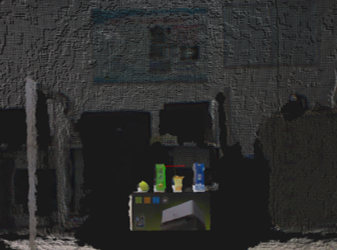}}
    \centerline{(a)}
  \end{minipage}
  \begin{minipage}{0.33\linewidth}
    \centerline{\includegraphics[width=1\textwidth]{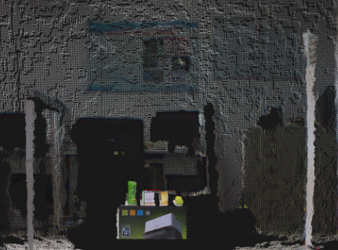}}
    \centerline{(b)}
  \end{minipage}
  \begin{minipage}{0.33\linewidth}
    \centerline{\includegraphics[width=1\textwidth]{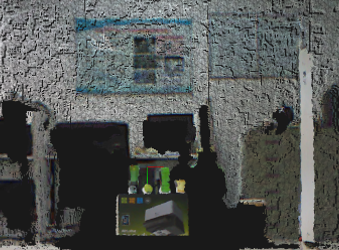}}
    \centerline{(c)}
  \end{minipage}
  \caption{Three samples of real scene point clouds: (a) There are ball, bottle1, cup, bottle2 on a planar object. (b) There are bottle1, box, cup, ball on a planar object. (c) There are bottle3, ball, bottle1, cup on a planar object.}
  \label{fig_real_scene}
  \end{figure*}

  We can obtain each individual target object from the real scene by utilizing the proposed method of segmentation. 
  Here, 150 different point clouds of the real scenes are used for this experiment. 
  Figure \ref{fig_real_scene} shows some samples of real scene point clouds. After that, we use the two aforementioned trained SVM classifiers to evaluate the performance of the Color-CVFH descriptor and the CVFH descriptor. 
  Table \ref{tab_7} and Table \ref{tab_8} show the results, from which we can observe that the classifier trained by the Color-CVFH outperform the classifier trained by the CVFH significantly.
  The classifier trained by the CVFH descriptor, is more likely to consider the bottle1 as bottle2 or bottle3. 
  Nearly one fifth of bottle1 are recognized as the category of others, and over half of the category of others (the bottle2 and bottle3 belong to the category of others) are classified as the category of bottle1. 
  Therefore, the classifier does not perform well for the category of bottle0 in terms of recall (${82.67}\%$) and precision (${61.39}\%$). 
  Because the cup and the box share some similarities in shape and color from a certain angle. 
  Some objects belonging to the cup are considered as the category of others (the box belong to the category of others). 
  As a result, the classifier does not have a perfect precision (${90}\%$) for the category of cup. 
  However, the classifier perform well for the category of ball in terms of recall, precision and F1\_score, since the ball is quite different from the other objects of this experiment in shape.

  \begin{table*}[htbp]
    \renewcommand{\arraystretch}{1.5}
    \centering
    \caption{The results of prediction for CVFH by testing in real scenes}
    \label{tab_7}
      \begin{tabular}{|c|c|c|c|c|}
      \hline
      \diagbox{Actual class}{Prediction class}&Bottle0&Cup&Ball&Others\cr
      \hline
      Bottle0&{\bf82.67\%}&0&0&17.33\%\cr\hline
      Cup &0&{\bf96\%}&0&4\%\cr\hline
      Ball&0&4\%&{\bf96\%}&0\cr\hline
      Others&52\%&6.67\%&0&{\bf41.33\%}\cr
      \hline
      \end{tabular}
  \end{table*}
  
  \begin{table*}[htbp]
    \renewcommand{\arraystretch}{1.5}
    \centering
    \caption{The results of the evaluation for CVFH and Color-CVFH by testing in real scenes}
    \label{tab_8}
      \begin{tabular}{|c|c|c|c|c|c|c|}
      \hline
      \multirow{2}{*}{}&
      \multicolumn{3}{c|}{CVFH}&\multicolumn{3}{c|}{Color-CVFH}\cr\cline{2-7}
      &Recall&Precision&F1\_score&Recall&Precision&F1\_score\cr
      \hline
      %\hline
      Bottle0 vs. all&82.67\%&61.39\%&0.70&{\bf 100\%}&{\bf 100\%}&{\bf 1}\cr\hline
      Cup vs. all&96.00\%&90.00\%&0.93&{\bf 100\%}&{\bf 100\%}&{\bf 1}\cr\hline
      Ball vs. all&96.00\%&100\%&0.98&{\bf 100\%}&{\bf 100\%}&{\bf 1}\cr\hline
      \end{tabular}
  \end{table*}
  
  In contrast, the classifier trained by the Color-CVFH obtains a perfect recognition performance in terms of recall, precision and F1\_score, 
  which can not only perform well on the objects with dissimilar shapes, but also on the objects with similar shapes and colors at category level, such as the bottle1 and bottle3 in this experiment. 
  Additionally, from the Table \ref{tab_8}, we can see that the classifier trained by Color-CVFH descriptor can classify each individual object from the the real scene correctly.
  
  The CVFH descriptor is based on the geometrical information of the 3D object surface only, thus it can obtain a good recognition performance on the objects with different shapes, like the plate in the first experiment. 
  But it is relatively ineffective in recognizing the objects which have similar shape information, like the apple2 and apple5 in the second experiment. 
  To cope with this problem, the proposed Color-CVFH descriptor combines the shape information with the color information, and the results of the above three experiments show that this proposed descriptor can acquire a much better recognition performance in both the public RGB-D image dataset and real scenes. 
  In addition, the proposed descriptor is effective at recognizing the objects which have similar shape information and color information at category level, like the object bottle1 and object bottle3 in the third experiment. 
  Though it makes some mistakes when recognizing the objects with almost the same shape and color at instance level, like the instance apple3 and instance apple4 in the second experiment, the proposed Color-CVFH descriptor can acquire a strikingly good recognition performance in most circumstance.

  \subsection{Testing the object recognition and grasping system}
  
\begin{figure}[htbp]
    \centering
    \includegraphics[width=0.48\textwidth]{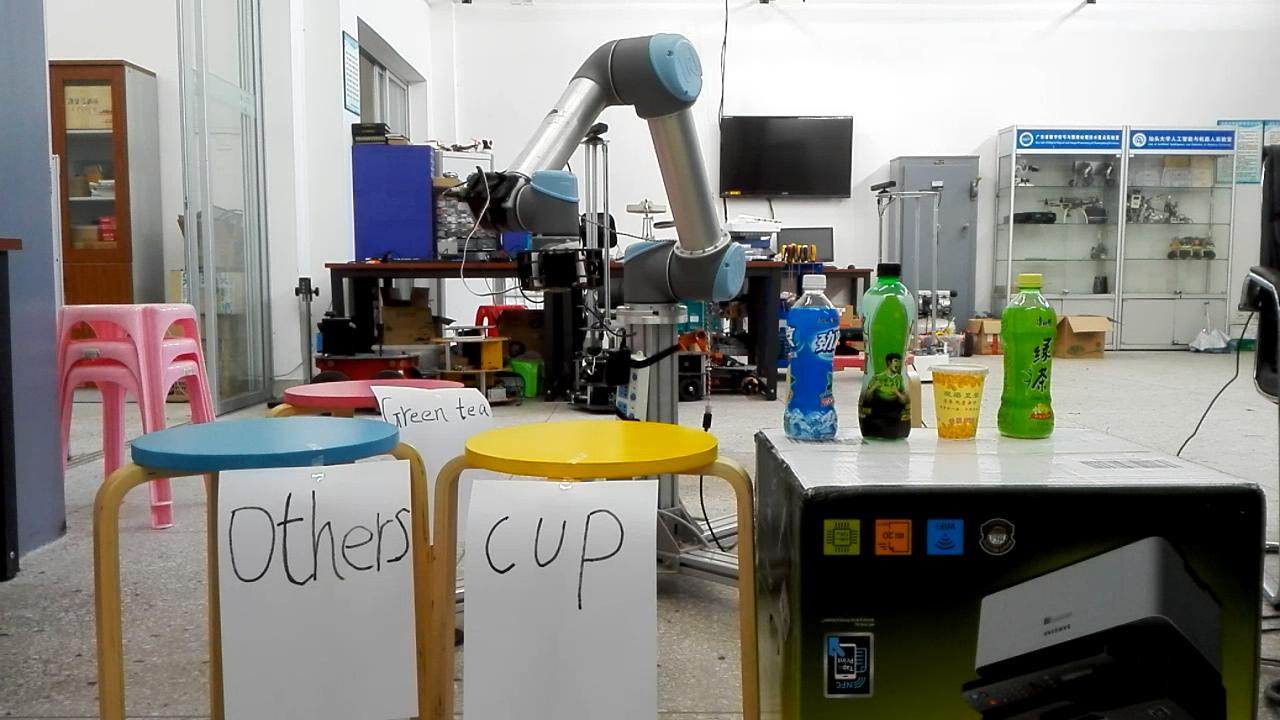}
    \caption{The real scene of the object recognition, grasping, and sorting task.}
    \label{fig_real_grasp}
  \end{figure}

  In this work, we utilize the four-classes SVM classifier trained by the proposed Color-CVFH descriptor in the third experiment to construct an object recognition, grasping, and sorting system. 
  Then we test its performance to recognize, grasp and sort four different objects, which are shown in Figure \ref{fig_real_grasp}.
  
  \begin{figure*}
    \begin{minipage}{0.33\linewidth}
      \centerline{\includegraphics[width=1.01\textwidth]{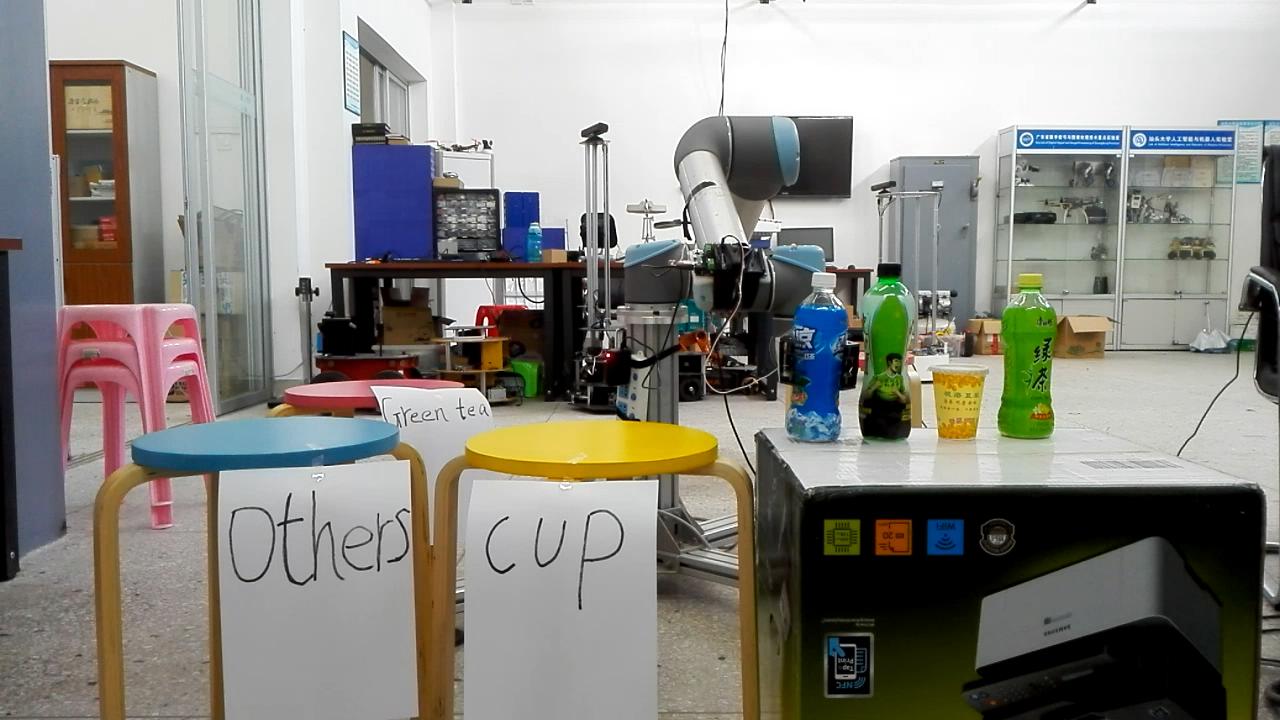}}
      \centerline{(a)}
    \end{minipage}
    \begin{minipage}{0.33\linewidth}
      \centerline{\includegraphics[width=1.01\textwidth]{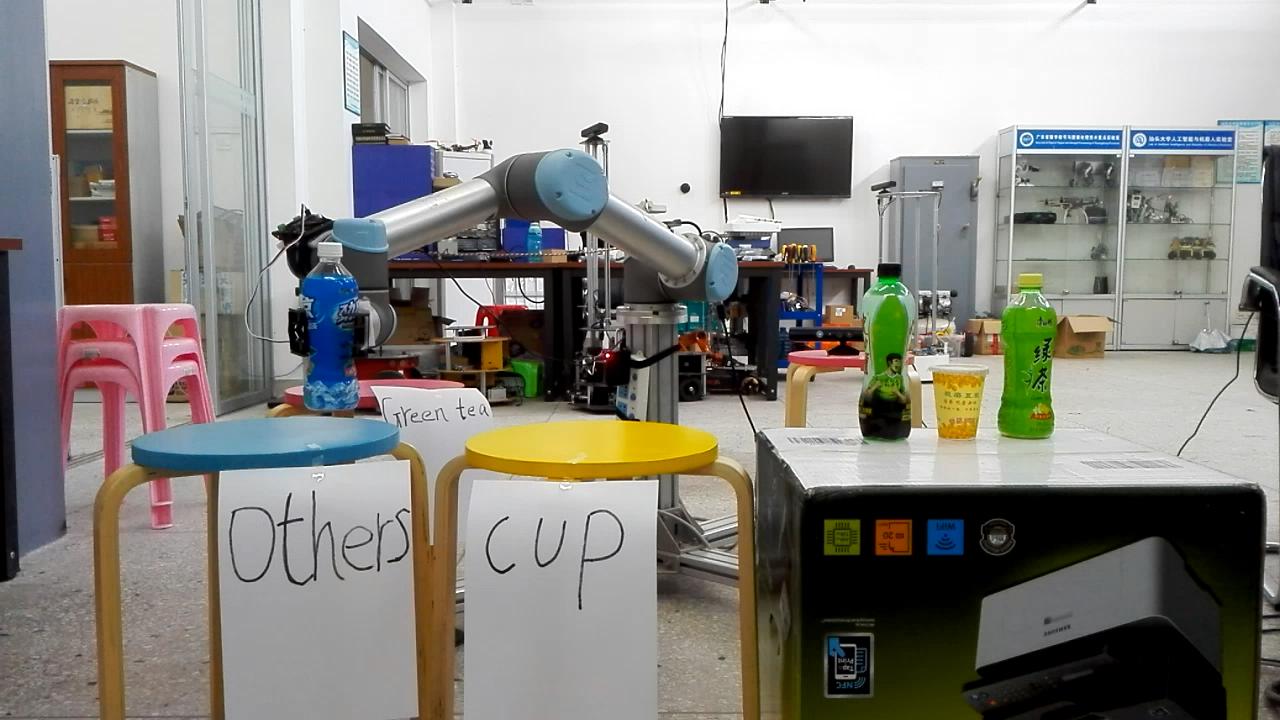}}
      \centerline{(b)}
    \end{minipage}
    \begin{minipage}{0.33\linewidth}
      \centerline{\includegraphics[width=1.01\textwidth]{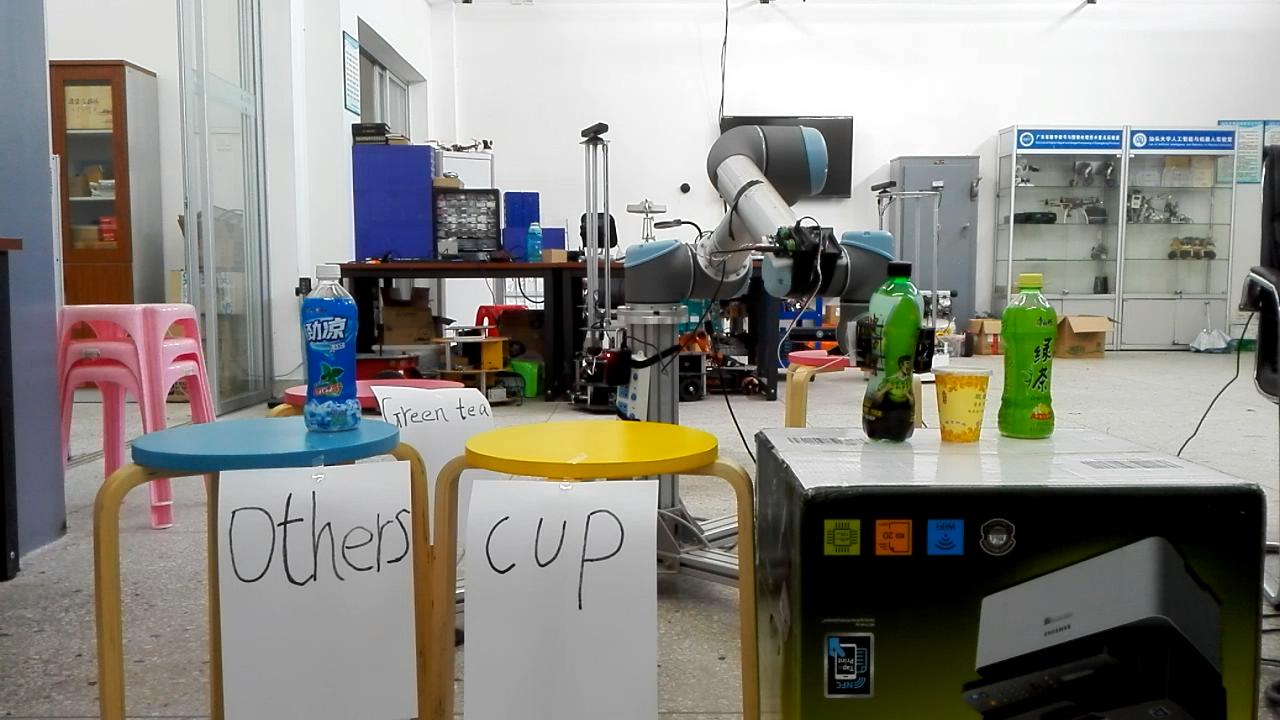}}
      \centerline{(c)}
    \end{minipage}
    \begin{minipage}{0.33\linewidth}
      \centerline{\includegraphics[width=1.01\textwidth]{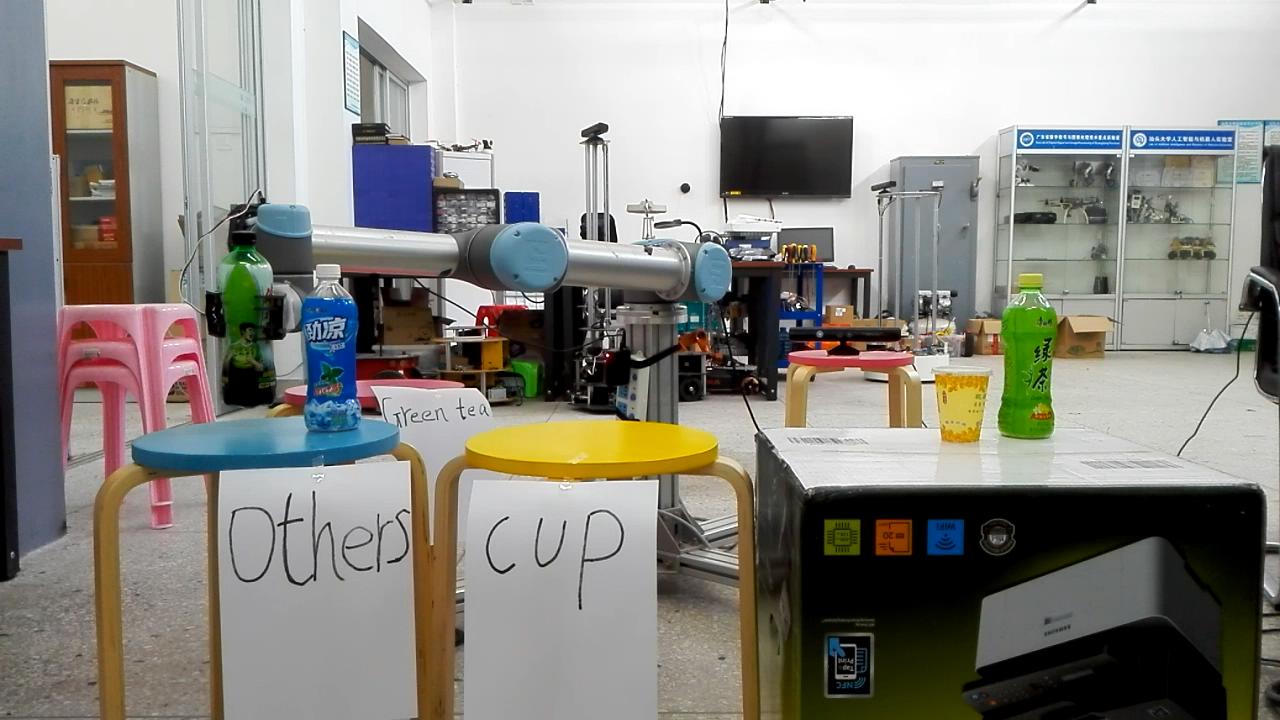}}
      \centerline{(d)}
    \end{minipage}
    \begin{minipage}{0.33\linewidth}
      \centerline{\includegraphics[width=1.01\textwidth]{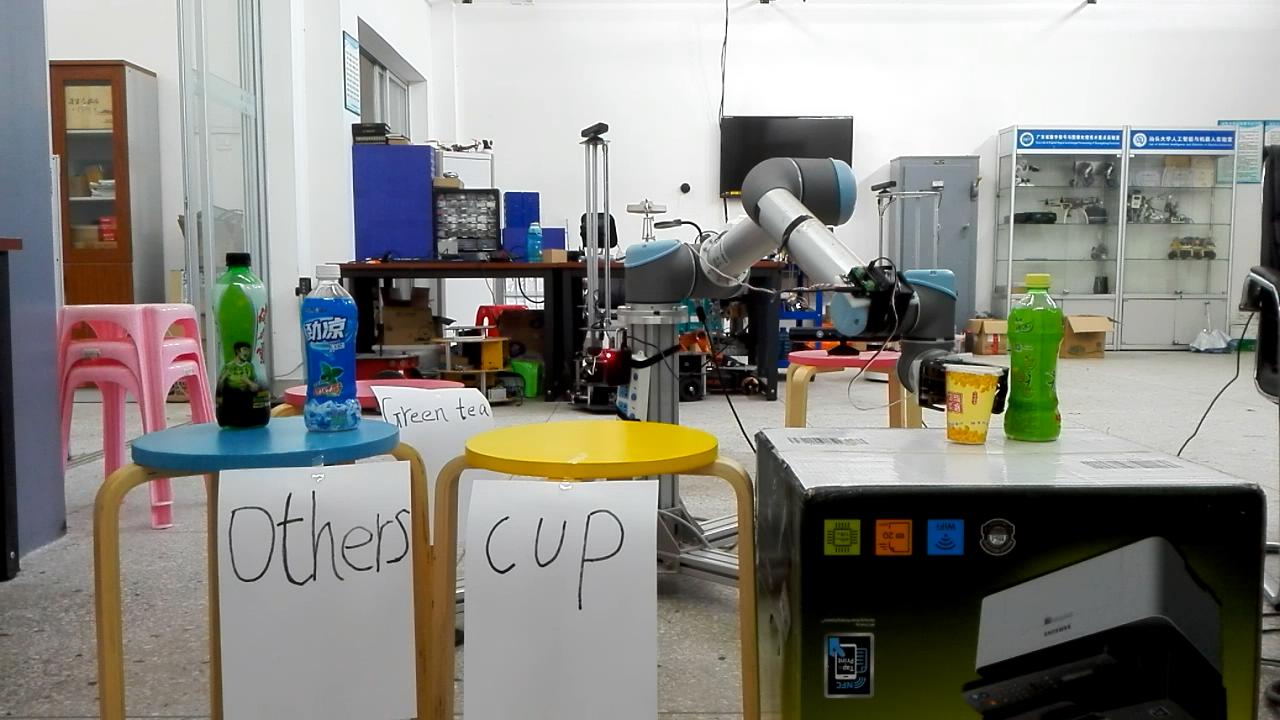}}
      \centerline{(e)}
    \end{minipage}
    \begin{minipage}{0.33\linewidth}
      \centerline{\includegraphics[width=1.01\textwidth]{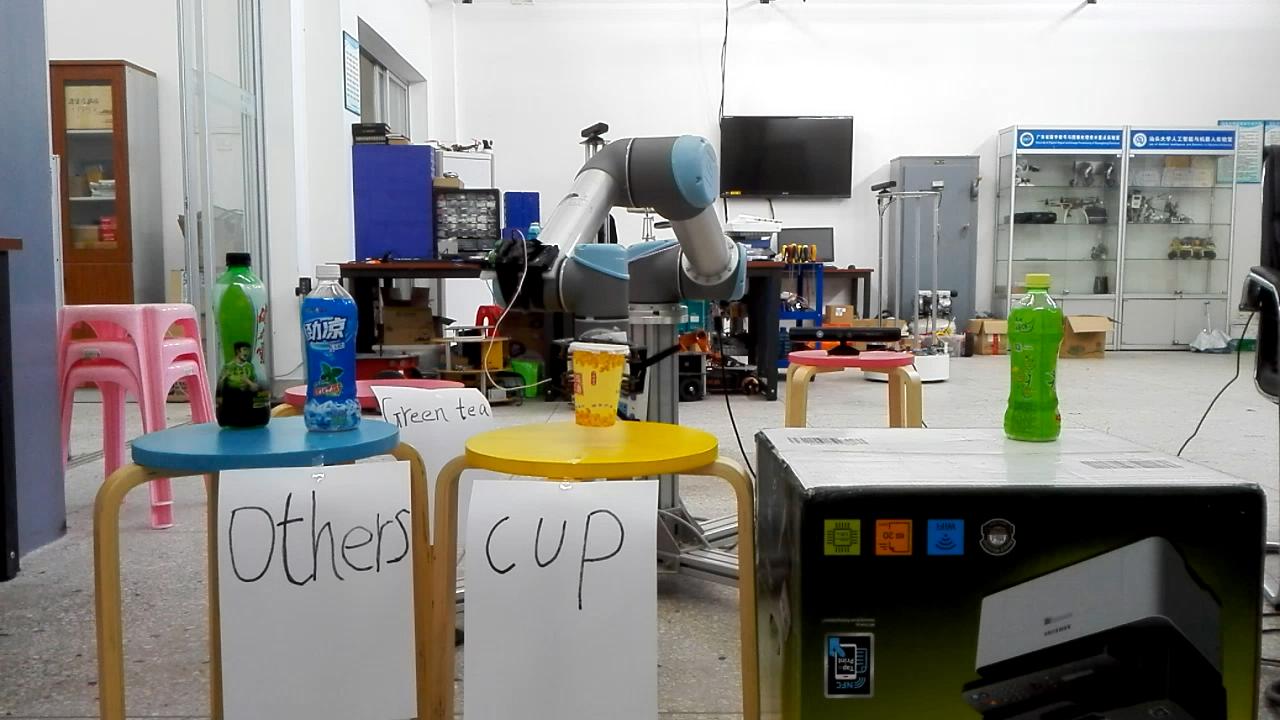}}
      \centerline{(f)}
    \end{minipage}
    \begin{minipage}{0.33\linewidth}
      \centerline{\includegraphics[width=1.01\textwidth]{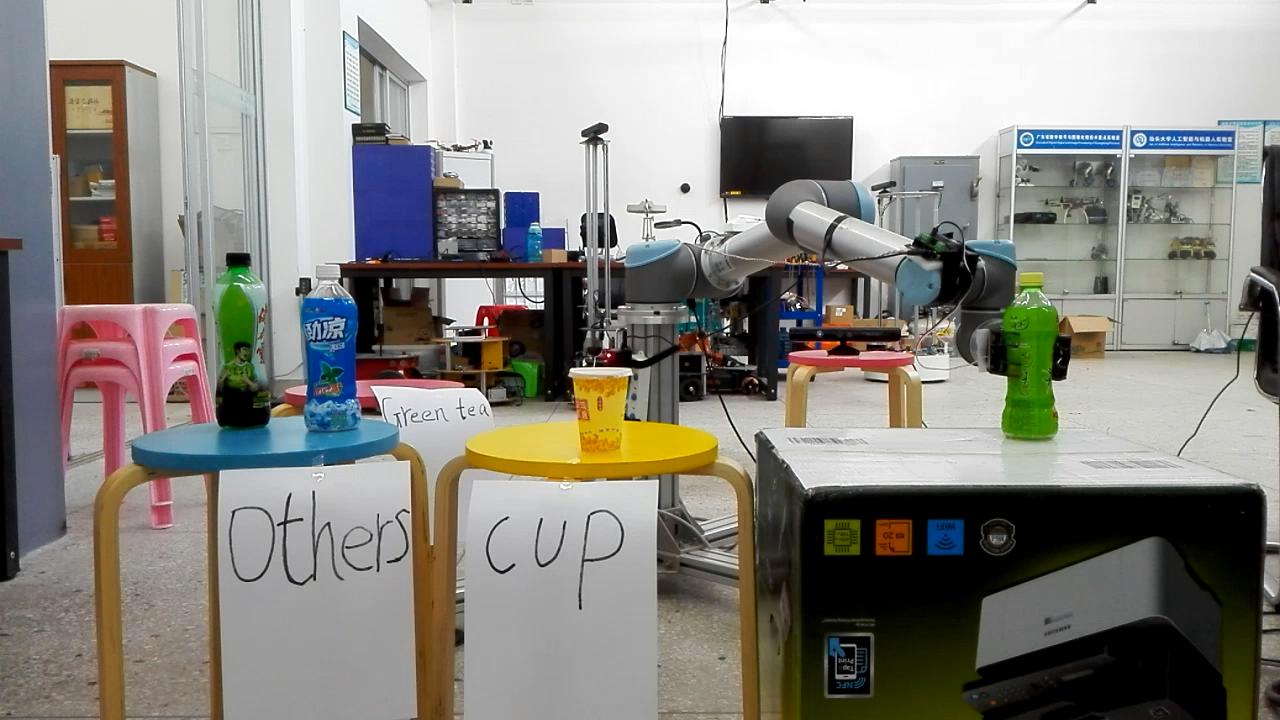}}
      \centerline{(g)}
    \end{minipage}
    \begin{minipage}{0.33\linewidth}
      \centerline{\includegraphics[width=1.01\textwidth]{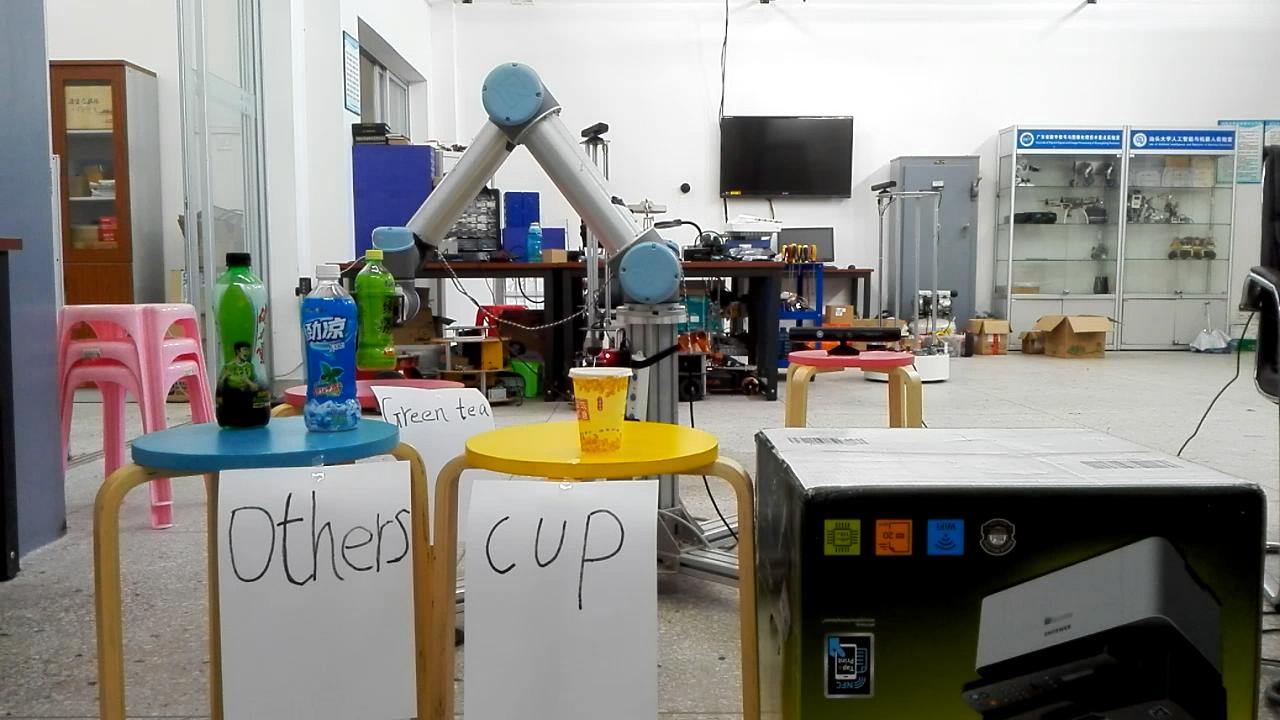}}
      \centerline{(h)}
    \end{minipage}
    \begin{minipage}{0.33\linewidth}
      \centerline{\includegraphics[width=1.01\textwidth]{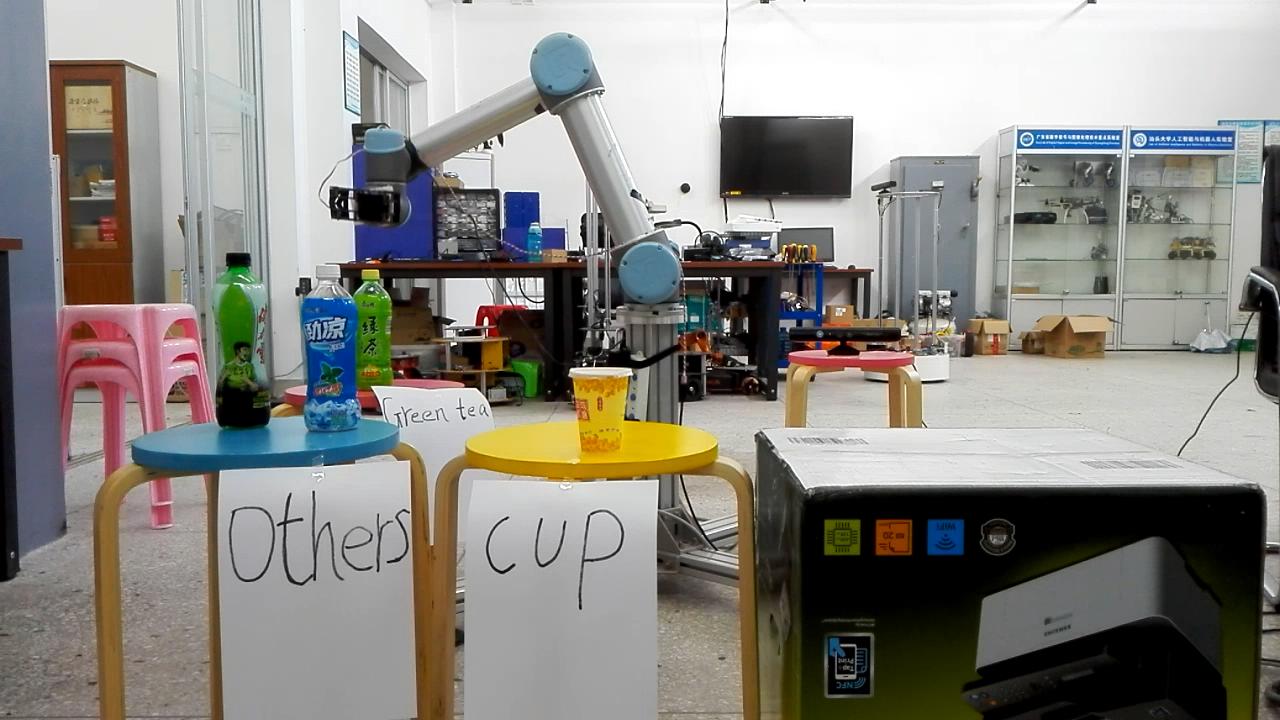}}
      \centerline{(i)}
    \end{minipage}
    \caption{The result of experiment for object recognition, grasping and sorting: (a) grasping the bottle2; (b) dropping the bottle2 to the category of others; (c) grasping the bottle3; 
    (d) dropping the bottle3 to the category of others; (e) grasping the cup; (f) dropping the cup to the category of cup; (g) grasping the bottle1; (h) dropping the bottle2 to the category of green tea; (i) completing sorting.}
    \label{fig_result}
    \end{figure*}
        
The Figure \ref{fig_result} shows that the proposed system can not only figure out the category of these four target objects, but also grasp the objects and place them to their specified locations successfully. 
For example, it can pick up the cup and place it on the stool with the label of cup. The experimental results show that the system can implement the object recognition, grasping and sorting tasks well.

\section{Conclusion and future work}
We present a global texture-shape 3D feature descriptor for object recognition in this paper, and utilize the multi-class SVM classifier for recognizing the target objects instead of using feature matching, which can reduce computing burden. 
Then we evaluate the recognition performance of the proposed descriptor with both public dataset and real scenes. 
The experimental results show that the classifier trained by the proposed descriptor outperforms the classifier trained by the CVFH descriptor for recognizing and classifying objects with similar shapes and colors. 
In the end, we present and test an object recognition, grasping, and sorting system, which employs the proposed Color-CVFH descriptor. 
The experimental results show that the proposed system can implement the object recognition, grasping and sorting tasks well.

The proposed 3D feature descriptor can perform very well at category level recognition, but it makes some mistakes when distinguishing the individual objects with almost the same shape and color information at instance level. 
Therefore, the pose estimation and more texture information will be considered to design more effective descriptors in the future work.

\section*{Acknowledgment}
This work was supported in part by the National Natural Science Foundation of China (NSFC) under grant 61300159, 61473241 and 61332002, 
by the Project of Internation as well as Hongkong, Macao \& Taiwan Science and Technology Cooperation Innovation Platform in Universities in Guangdong Province under grant 2015KGJH2014, 
by the Science and Technology Planning Project of Guangdong Province of China under grant 2013B011304002, by Educational Commission of Guangdong Province of China under grant 2015KGJHZ014.

\bibliographystyle{SageV}
\bibliography{Object_sorting.bib}

\end{document}